\documentclass{IEEEtran}
\usepackage{amsmath}
\usepackage{multicol}
\usepackage{geometry}
\usepackage{graphicx}

\begin{document}
\title{Deep Metric Learning with Locality Sensitive Angular Loss for Self-Correcting Source Separation of Neural Spiking Signals}
\author{Alexander Kenneth Clarke,  Dario~Farina,~\IEEEmembership{Fellow,~IEEE}
\thanks{D. Farina (d.farina@imperial.ac.uk) and A. K. Clarke (a.clarke18@imperial.ac.uk) are with the Department of Bioengineering, Imperial College London, UK.}
\thanks{The work was supported by the EPSRC Centre for Neurotechnology 	(EP/L016737/1) (AKC) and the EPSRC Transformative Healthcare Technologies 2050 Grant (EP/T020970/1) (DF).}
}
\thanks{This work has been submitted to the IEEE for possible publication. Copyright may be transferred without notice, after which this version may no longer be accessible.
}

\maketitle
\begin{abstract}
Neurophysiological time series, such as electromyographic signal and intracortical recordings, are typically composed of many individual spiking sources, the recovery of which can give fundamental insights into the biological system of interest or provide neural information for man-machine interfaces. For this reason, source separation algorithms have become an increasingly important tool in neuroscience and neuroengineering. However, in noisy or highly multivariate recordings these decomposition techniques often make a large number of errors, which degrades human-machine interfacing applications and often requires costly post-hoc manual cleaning of the output label set of spike timestamps. To address both the need for automated post-hoc cleaning and robust separation filters we propose a methodology based on deep metric learning, using a novel loss function which maintains intra-class variance, creating a rich embedding space suitable for both label cleaning and the discovery of new activations. We then validate this method with an artificially corrupted label set based on source-separated high-density surface electromyography recordings, recovering the original timestamps even in extreme degrees of feature and class-dependent label noise. This approach enables a neural network to learn to accurately decode neurophysiological time series using any imperfect method of labelling the signal.
\end{abstract}


\section{Introduction}
\IEEEPARstart{T}{ime} series neurophysiological data is commonly characterised by repetitive activation events, for example the motor unit activation potentials (MUAP) in electromyographic (EMG) signals or the spike potentials in microelectrode cortical recordings\cite{merletti2016surface}\cite{Stark2007}. The ensemble of these activation events constitute neural codes that give a direct insight into the target system\cite{Drebitz2019}, whilst also providing an accurate control signal for human-machine interfacing applications, such as prosthetic control\cite{Farina2016}\cite{Kapelner2019}. Neurophysiological signals are typically linear superpositions of many of these spiking sources, and their extraction from noisy systems has long been a major focus of applied source separation in neuroscience\cite{Calvin1973}. 

Identifying sources from multiunit activity was originally achieved through manual sorting by trained operators\cite{Simon1965}, but this tedious process was quickly supplanted by automated methods using early forms of blind source separation (BSS)\cite{Rey2015}. BSS algorithms have since become extremely effective, able to automate the recovery of sources in highly noisy and complex systems\cite{Rey2015}\cite{Farina2016}\cite{Kevric2017}. An important trend within the field of applied source separation has been the increasing availability of highly multivariate data\cite{Negro2016}\cite{Pachitariu2016}, as a result developments in high-density electrode arrays\cite{Steinmetz2017}\cite{Muceli2019}. By exploiting the increased spatial information collected by these systems, BSS pipelines can yield extremely large numbers of sources\cite{Pachitariu2016}\cite{DelVecchio2019b}. A more recent development is the adoption of deep learning approaches as a replacement for linear separation vectors, using neural networks to decode signals with a high degree of robustness to noise and signal non-stationarities\cite{Clarke2021}\cite{Wen2021}. These methods involve an offline supervised training phase using the augmented output of a BSS algorithm, therefore an important requirement is that the BSS decomposition contains relatively few errors if the network is to decode with high accuracy\cite{Clarke2021}. 

As the number of sources identified in a manual or automatic decomposition increases, so does the probability of labelling errors. Noise can be mistakenly labelled as an activation, an activation can be assigned to the wrong class or missed entirely, a class might be inappropriately partitioned or two distinct classes merged into one. These sources of label noise can be categorised based on whether the factors affecting the likelihood of a label from one class to flip to that of a different class are shared at the dataset, class or feature level\cite{algan2020label}. As label-flipping is class or even feature-dependent, it is difficult to identify such errors automatically, and for automatic decompositions a degree of manual post-hoc cleaning is commonly employed, often using additional knowledge about the system of interest, such as temporal statistics in the source activations\cite{Kumar2020}. The nature of this manual cleaning generally relates to the mixing system of interest, for example intracortical and intramuscular EMG (iEMG) decompositions generally require post-hoc examination of classes due to extensive class-dependent label noise\cite{Rey2015}\cite{McGill2005}. On the other hand, surface EMG (sEMG) decompositions also contain a degree of feature-dependent noise and so require further inspection of specific activations\cite{Hug2021}. Whilst accurate, manual post-hoc "cleaning" is an extremely time-consuming process and in some cases not feasible because of the size of the datasets being source-separated\cite{Carlson2019}. For this reason, modern source separation pipelines are increasingly using additional automated post-processing steps in an attempt to reduce the false label burden\cite{Kumar2020}\cite{Yger2018}\cite{Negro2016}. However, these methods only compensate for a relatively small proportion of incorrect labels, so that there remains a need for new methods of post-hoc label cleaning. Additionally, if supervised deep learning frameworks are to be trained using BSS-labelled signals with increasing degrees of label noise, then they need to be able to detect and manage such errors, i.e. they need to be designed to be implicitly self-correcting.

Managing label noise is a fast-developing subfield of modern machine learning. As the size of datasets expand faster than the capacity of domain experts to screen and label, data scientists are increasingly turning to new labelling methods that are more scalable at the cost of a greater proportion of label noise, such as Amazon’s Mechanical Turk\cite{song2020learning}. This is particularly true in a neuroscience setting, where datasets are generally labelled by a small pool of domain experts who can differ in professional opinion\cite{Karimi2019}. Approaches to learning in the presence of label noise can be broadly split into two categories; methods that aim to select models that are robust to label noise and methods that attempt to clean the label set prior to training\cite{Frenay2014}. Contemporary methods based on the latter approach generally rely on additional models which attempt to identify noisy labels using either a smaller pool of known correct labels or by comparing a label with other in-class labels using a similarity metric\cite{VeitACKGB17}\cite{DBLP:Lee}\cite{DBLP:Han}. This principle of using similarity metrics to build embedding spaces that inform intra- and inter-class classification is closely related to deep metric learning (DML) approaches.

The objective of deep metric learning is the training of a deep neural network which maps discrete inputs to an embedding space in which positive pairs (two inputs from the same class) are closer than negative pairs (two inputs from different classes)\cite{Chopra}. The Euclidean distance is commonly used as the distance metric, although measures based on the angular difference between embeddings have become popular due to their inherent rotation and scale invariance\cite{wang2017deep}. The network is optimised either using the absolute similarity between pairs or, more commonly, the relative difference between one or more positive and negative pairings\cite{Schroff_2015_CVPR}. During optimisation, most negative pairs will quite quickly become much further away than positive pairs, thus random generation of pairs for training is extremely inefficient. For this reason many implementations of DML seek to select the most informative pairings from each minibatch by applying some form of selection rule operating on the output embedding space\cite{HermansBL17}\cite{NIPS2016_Sohn}. Most of these approaches are designed purely to maximise the class separability of the embedding space, leading to dense clusters in each class\cite{Wu_2017_ICCV}. More recent work has attempted to increase intra-class variability, as such tight embeddings potentially reduce the ability of the model to generalise to new in-class inputs\cite{rippel2016metric}\cite{Wu_2017_ICCV}\cite{Wang_2019_CVPR}. These locality-sensitive approaches have a less distorting effect on the embedding space, giving better generalisation performance\cite{Wang_2019_CVPR}. Whilst not the primary goal of these studies, another effect of preserving intra-class variance is a richer embedding of inputs, with semantically similar events sub-clustering\cite{rippel2016metric}. Such rich embedding spaces could potentially be of use for detecting class outliers, perhaps even having utility for label-cleaning operations in event-driven neurophysiological recordings. DML pipelines have been designed to operate on noisy label sets for similar tasks such as person reidentification, however these methods tend to use an external method to modify training rather than the embedding space itself, for example using label-correction based on cross-entropy\cite{Xu2021}.  

Motivated by the need to further preserve a rich intra-class embedding, and inspired by ranking approaches to triplet sampling, such as\cite{Wang_2019_CVPR}, here we propose a novel locality-sensitive approach to sampling during DML optimisation. We use an efficient top-k query to identify the closest in-batch positive to each event and the N-closest in-batch negatives, which are then used to calculate an N-pair formulation of the popular angular loss\cite{wang2017deep}. The idea of using top-k queries within batch losses has been explored in the context of binary classification problems\cite{fan2017learning}, however this is, to our knowledge, the first such implementation within the domain of DML. In this paper we demonstrate that this simple modification, which we call locality-sensitive angular loss (LSAL), generates an embedding space which can be used to detect and classify repetitive events, whilst importantly having the additional utility of being able to detect label-noise in the data used for training. 

The main contribution of this paper is DeepLSAL, a novel DML pipeline that leverages LSAL to perform both label-cleaning and the identification of new activations in unseen data, operating directly on neurophysiological time series signals. The specific focus is on label sets generated by BSS algorithms for decoding convolutive mixtures, as this is an area where supervised deep learning methods are clearly beginning to outperform existing methods\cite{Clarke2021}\cite{Wen2021}; however, in principle the proposed methodology is applicable to any manual or automatic method of generating imperfect label sets. The effectiveness of DeepLSAL is validated robustly using an experimentally-collected high-density (HD) sEMG dataset that had been source-separated into constituent motor unit activity using the gradient convolutional kernel compensation (gCKC) algorithm\cite{holobar2007multichannel}. The scenario of decomposition of HD-sEMG signals is highly convenient for validation of the proposed approach since both the generating system and decomposition methodology are well-studied. Moreover, the sEMG system is characterised by a high degree of feature-dependent label noise due to the complex superposition of MUAPs caused by volume-conduction effects\cite{Farina2016a}. 

\section{Theory and Algorithm}
\subsection{Deep Metric Learning with N-pair Loss}
The objective of ranking loss DML, also called triplet loss, is to train a deep learning function such as a convolutional neural network to map a sample taken from one of \begin{math}C\end{math} classes to an embedding vector \begin{math}\mathbf{x}\end{math}, such that for an arbitrarily selected anchor embedding \begin{math}\mathbf{x_{a}}\end{math}, the embedding space reduces the relative distance \begin{math}\mathbf{D}\end{math} between positive samples from the same class \begin{math}\mathbf{x_{p}}\end{math} and negative samples from different classes \begin{math}\mathbf{x_{n}}\end{math}. \begin{math}\mathbf{D}\end{math} can be a number of different metrics, such as the Euclidean distance, cosine similarity or Kullback–Leibler divergence\cite{Ji2020}. Commonly the loss function is formulated such that the relative difference be greater than a margin \begin{math}m\end{math} such that:

\begin{equation}
\small
\mathbf{D}(\mathbf{x_{a}},\mathbf{x_{n}}) + m \leq \mathbf{D}(\mathbf{x_{a}},\mathbf{x_{p}})
\label{equation:1}
\end{equation}

\begin{figure*}
\centering
\includegraphics[width=1\textwidth]{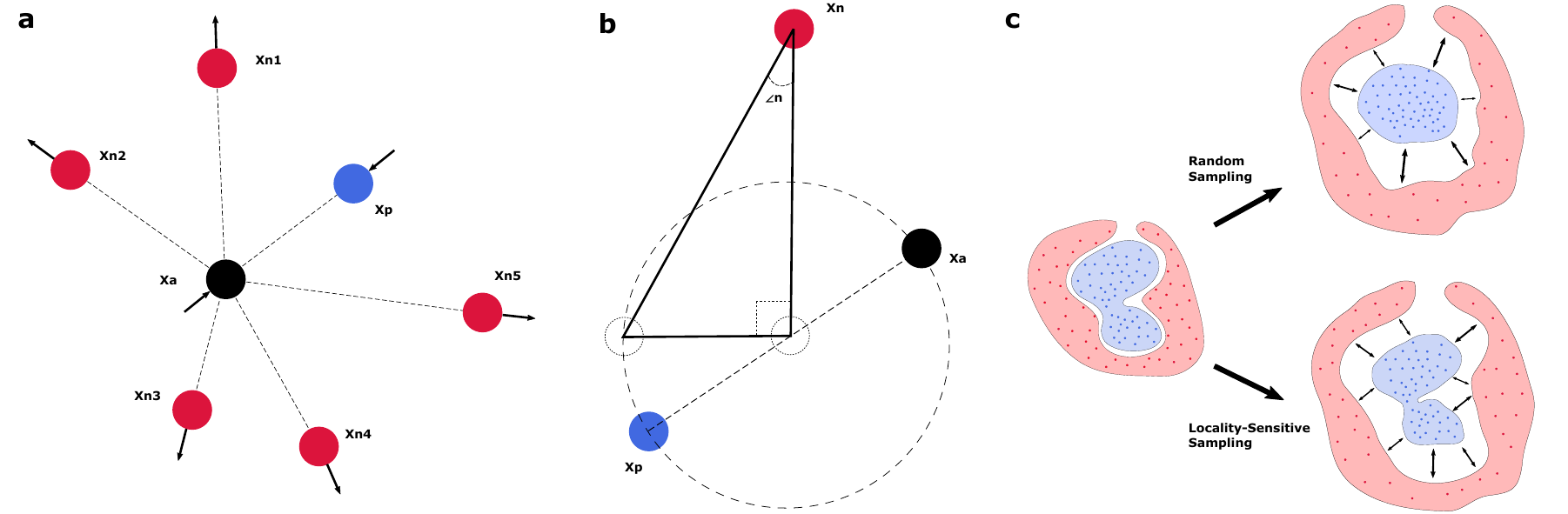}
\caption{\textbf{a} shows the principle behind N-pair loss, in this case with an N of 5. As multiple negative embeddings are utilised, the chance of sampling only uninformative pairings reduces. \textbf{b} illustrates the principle behind vertex selection in angular loss. Rather than directly using the triplet as vertices, only the negative embedding is used, whilst the other two vertices are constructed so as to build a right-angled triangle with the 90\textdegree angle at the midpoint between the anchor and positive embedding. \textbf{c} demonstrates the effect of random versus locality-sensitive sampling on the intra-class variance.}
\label{figure:1}
\end{figure*}
 
After a small amount of optimisation, the bulk of negative pairs will be much further away than the positive pairs, meaning most training examples in a batch will become uninformative\cite{HermansBL17}. N-pair loss seeks to avoid this problem by comparing each positive pair in an \begin{math}M\end{math}-size batch \begin{math}B\end{math} to \begin{math}N\end{math} multiple negative pairs (fig. \ref{figure:1}a), which are then combined in a \begin{math}log-sum-exp\end{math} formulation\cite{NIPS2016_Sohn}:

\begin{equation}
\small
L = \frac{1}{M}\sum_{\mathbf{x}_{a}\in B}^{M}log\left [1 + \sum_{\mathbf{x}_{n}\in B\setminus(\mathbf{x}_{a},\mathbf{x}_{p})}^{N}\exp(f_{a,p,n})\right]
\label{equation:2}
\end{equation}

where \begin{math}f_{a,p,n}\end{math} is usually a hinge function such as \begin{math}max(0, \mathbf{D}(\mathbf{x_{a}},\mathbf{x_{n}}) - \mathbf{D}(\mathbf{x_{a}},\mathbf{x_{p}}) + m)\end{math}. By taking an average across the negative pairs, it is likely that at least some informative negative pairings will be included in the loss calculation.

\subsection{Angular Loss}
Angular loss is based on a geometric reformulation of the distance metric; rather than minimising the distance of \begin{math}\mathbf{x_{p}}\end{math} to \begin{math}\mathbf{x_{a}}\end{math} relative to \begin{math}\mathbf{x_{n}}\end{math}, it instead minimises the angle at \begin{math}\mathbf{x_{n}}\end{math} of a triangle formed from the three embeddings. This has the effect of improving optimisation stability as angles are scale invariant, whilst using a triangle means all edges of the triplet are taken into account\cite{wang2017deep}. However, in certain circumstances the minimisation of the angle at \begin{math}\angle\mathbf{x_{n}}\end{math} will push \begin{math}\mathbf{x_{n}}\end{math} towards \begin{math}\mathbf{x_{a}}\end{math}. This can be avoided by constructing a right-angled triangle with \begin{math}\mathbf{x_{n}}\end{math} and the midpoint between \begin{math}\mathbf{x_{a}}\end{math} and \begin{math}\mathbf{x_{p}}\end{math} (fig. \ref{figure:1}b), with the final vertex being the point on the semicircle joining \begin{math}\mathbf{x_{a}}\end{math} and \begin{math}\mathbf{x_{p}}\end{math} which creates a right-angled triangle\cite{wang2017deep}. By dropping constant terms, this geometric relationship can be used for the \begin{math}f_{a,p,n}\end{math} in equation 2, expressed as:

\begin{equation}
\small
f_{a,p,n} = 4tan^{2}\alpha(\mathbf{x}_{a} + \mathbf{x}_{p})^{T}\mathbf{x}_{n} -2(1+tan^{2}\alpha)\mathbf{x}_{a}^{T}\mathbf{x}_{p}
\label{equation:3}
\end{equation}

where \begin{math}\alpha \end{math} is an angle in radians which sets the upper accepted bounds of the loss, analogous to \begin{math}m\end{math} in equation 1.

\subsection{Inducing Rich Embeddings}
Whilst a random selection of positive pairings in \begin{math}B_{ang}\end{math} is suitable if the objective is to maximise inter-class distance within the embedding space, it also has the tendency to collapse intra-class distances down to a point, as illustrated in figure \ref{figure:1}c\cite{rippel2016metric}. We theorised that the main explanation for this compression is that the sampling process is random, meaning that the neural network is induced during training to bring all embeddings from the same class together, which, as a complex non-linear function, it is quite capable of doing given sufficient training steps. We instead elected to sample positive and negative pairs based on the local neighbourhood of each embedding in the batch, modifying the selection of \begin{math}\mathbf{x_{p}}\end{math} and the set of \begin{math}\mathbf{x_{n}}\end{math}'s for each \begin{math}\mathbf{x_{a}}\end{math} in the batch \begin{math}B\end{math}, a process we term locality-sensitive sampling. 

\begin{math}B\end{math} is first selected such that it be large enough to have a diverse representation of each class. As each embedding vector is L2-normalised, the tensor formed by finding the inner product of the batch tensor with its transpose is the pairwise cosine similarity. For each \begin{math}\mathbf{x_{a}}\end{math} the \begin{math}\mathbf{x_{p}}\end{math} selected with a simple argmax, i.e. the closest different vector from the same class is selected. A similar procedure is used to select the set of \begin{math}\mathbf{x_{n}}\end{math}'s, using a top-k algorithm to select the \begin{math}k\end{math} most similar vectors to \begin{math}\mathbf{x_{a}}\end{math} that belong to a different class. GPU implementations of top-k algorithms have become extremely efficient in recent years, due to their increasing use within machine learning paradigms\cite{ShazeerMMDLHD17}. With these easily-implemented changes the N-pair formulation of the angular loss becomes:

\begin{equation}
\small
\begin{aligned}
l_{LSAL} ={} & \frac{1}{M}\sum_{\mathbf{x}_{a}\in B_{r}}^{M}\:\sum_{\mathbf{x}_{p}\in B\setminus(\mathbf{x}_{a},\mathbf{x}_{n})}\mathbf{1}_{\mathbf{x}_{p}\in \Upsilon }\:log\left[1 + g\right]\\
g ={} &\sum_{\mathbf{x}_{n}\in B\setminus(\mathbf{x}_{a},\mathbf{x}_{p})}\ \mathbf{1}_{\mathbf{x}_{p}\in \Theta} \: exp(f_{a,p,n})
\end{aligned}
\label{equation:4}
\end{equation}

where \begin{math}\Upsilon = argmax(\angle (\mathbf{x}_{a},\mathbf{x}_{p}))\end{math} is the argmax set of the pairwise cosine similarity \begin{math}\angle\end{math} between \begin{math}\mathbf{x_{a}}\end{math} and its associated set \begin{math}\mathbf{x_{p}}\end{math} in \begin{math}B\end{math} and \begin{math}\Theta = topk(\angle (\mathbf{x}_{a},\mathbf{x}_{n}))\end{math} is the top-k values of the ordered pairwise cosine similarity \begin{math}\angle\end{math} between \begin{math}\mathbf{x_{a}}\end{math} and its associated set \begin{math}\mathbf{x_{n}}\end{math} in \begin{math}B\end{math}. \begin{math}\mathbf{1}\end{math} is the indicator function.

To stabilise early optimisation we combined the LSAL loss with a categorical cross-entropy with temperature given by:

\begin{equation}
\small
l_{cross} = -\frac{1}{M}\sum_{i}^{M}\sum_{c}^{C}\mathbf{1}_{y_{i}\in C_{c}}log\frac{exp\left [  \frac{1}{\gamma}\mathbf{z}_{i}\right ]}{\sum_{j}^{C}exp\left [\frac{1}{\gamma}\mathbf{z}_{j}\right ]}
\label{equation:5}
\end{equation}

where \begin{math}\mathbf{z} = \mathbf{\frac{W}{\left \| W \right \|}}_{2}\mathbf{x}\end{math} and \begin{math}\mathbf{W}\end{math} is a trainable matrix that compresses the embedding vector down to a dimension C vector for comparison with the one-hot encoded class labels. This gives the final loss function:

\begin{equation}
\small
l = l_{LSAL} + \tau l_{cross}
\label{equation:6}
\end{equation}

\subsection{Source Separation}
The timestamps used by DeepLSAL can be generated using a wide variety of manual and automated processes, however for this study the gradient convolution kernel compensation (gCKC) algorithm was selected due to its strong performance in HD-sEMG signal decomposition\cite{Holobar}\cite{holobar2007multichannel}. In the gCKC framework for blind source separation, the vector of spiking sources \begin{math}s\end{math} at time \begin{math}t\end{math} are first extended with \begin{math}L\end{math} delayed versions of themselves, allowing the mixing problem, which is convolutive in most neurophysiological settings, to be written in instantaneous form: 

\begin{equation}
\small
x(t) = \mathbf{H}\tilde{s}(t-l) + \omega(t)
\label{equation:7}
\end{equation}

where the signal observation vector \begin{math}x\end{math} at time \begin{math}t\end{math} is a linear mixture parameterised by the operation of the mixing matrix \begin{math}\mathbf{H}\end{math} on the extended source vector \begin{math}\tilde{s}\end{math} plus noise \begin{math}\omega\end{math}. In practice both the observation and source vectors are additionally extended with a further \begin{math}R\end{math} values for reason of numerical stability during the source separation procedure. 

Unlike independent component analysis methods which seek to directly estimate a separation vector for each source, gCKC seeks to include the additional statistical information that the spiking sources generate repetitive events within the signal. Sources are instead estimated indirectly using a linear minimum mean square error estimator, with the estimated \begin{math}j\end{math}th source \begin{math}\hat{s}_{j}\end{math} at time point \begin{math}t\end{math} given by:

\begin{equation}
\small
\hat{s}_{j}(t) = \hat{c}_{\mathbf{s}_{j}\tilde{\mathbf{x}}}^{T} \mathbf{C}_{\tilde{\mathbf{x}}\tilde{\mathbf{x}}}^{-1}\tilde{x}(t)
\label{equation:8}
\end{equation}

where \begin{math}\hat{\mathbf{c}}_{\hat{{\mathbf{s}}_{j}\tilde{\mathbf{x}}}}^{T}\end{math} is the transposed cross-correlation vector between an activation of the \begin{math}j\end{math}th source and extended HD-sEMG matrix and 
\begin{math}\mathbf{C}_{\tilde{\mathbf{x}}\tilde{\mathbf{x}}}^{-1}\end{math} is the inverted autocorrelation matrix of the extended HD-sEMG matrix \begin{math}\tilde{\mathbf{x}}\end{math}.

The vector \begin{math}\hat{c}_{\hat{{\mathbf{s}}_{j}\tilde{\mathbf{x}}}}^{T}\end{math} is usually initialised with a time point likely to contain a source activation, which can be estimated by, for example, the Mahalanobis distance calculated on the signal\cite{holobar2007multichannel}. Once selected,  \begin{math}\hat{\mathbf{c}}_{\hat{{\mathbf{s}}_{j}\tilde{\mathbf{x}}}}^{T}\end{math} is then optimised to find the rest of the source's signal contributions. This can be done with either a fixed-point algorithm as in \cite{Negro2016} or in the gCKC formulation by gradient descent:

\begin{equation}
\small
\mathbf{c}_{\hat{\mathbf{s}}_{j}\tilde{\mathbf{x}}}^{\circ} = \mathbf{c}_{\hat{\mathbf{s}}_{j}\tilde{\mathbf{x}}} - \alpha\sum_{t}\frac{\partial f\big(\hat{\mathbf{s}}_{j}\big(t\big)\big)}{\partial \hat{\mathbf{s}}_{j}\big(t\big)}\tilde{\mathbf{x}}\big(t\big)
\label{equation:9}
\end{equation}

where \begin{math}\mathbf{c}_{\hat{\mathbf{s}}_{j}\tilde{\mathbf{x}}}^{\circ}\end{math} is the updated cross-correlation vector, \begin{math}\alpha\end{math} is the learning rate and \begin{math}f\big(.\big)\end{math} a contrast function designed to estimate the non-gaussianity of the output source in a similar fashion to independent component analysis. Optimised sources can then be converted to timestamps using a linear threshold or a two-class k means clustering algorithm.  

\begin{figure*}
\centering
\includegraphics[width=1\textwidth]{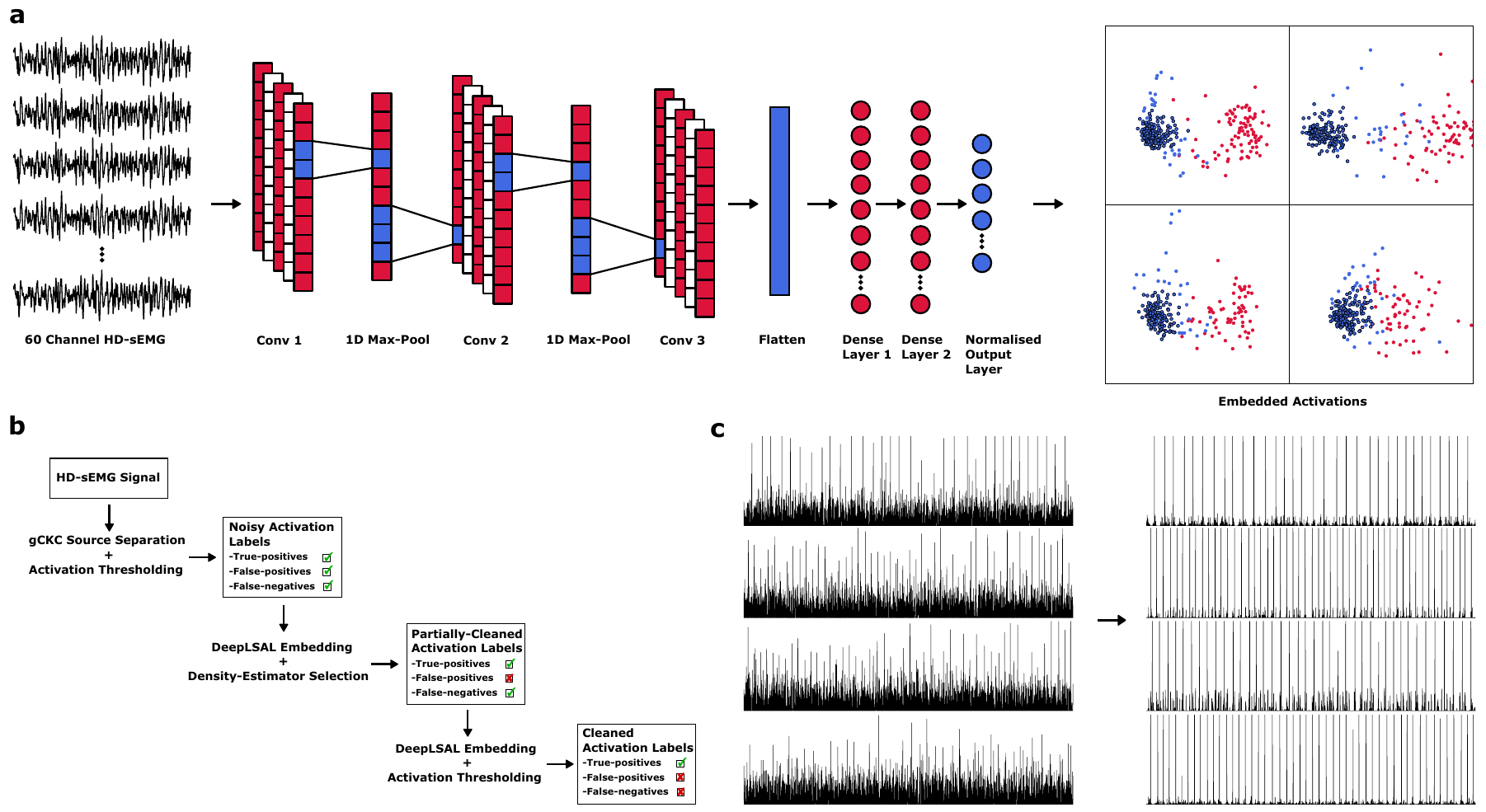}
\caption{\textbf{a} shows the model used in DeepLSAL, which takes the form of an easily-implemented convolutional neural network trained using a locality-sensitive angular loss to embed windows of neurophysiological time series into a low-dimensional space which can be used to both source separate and label clean. \textbf{b} shows the full DeepLSAL pipeline by which the noisy activation labels found from source-separating the high-density surface electromyography signal are cleaned. DeepLSAL is run twice - a cleaning phase to find the false positive labels and and a refitting phase to find the false negatives. After the refitting phase the predicted class activity is much cleaner than that of the original source separation algorithm, as seen in \textbf{c}.}
\label{figure:2}
\end{figure*}

\section{Validation Methodology}
\subsection{Experimental Dataset}
The HD-sEMG data set consisted of a set of 20-second recordings taken from the dominant tibialis anterior muscle of 10 men performing an isometric contraction at 15\% of maximal force, previously used to validate source separation techniques\cite{holobar2010experimental}. Maximal contraction was defined as the mean force of three 5-s maximal contractions separated by 3 min of rest, with force sampled at 2048Hz by load cells mounted on an isometric brace. Force feedback was provided to the participants by an oscilloscope. The signal from a monopolar \begin{math}12\times 5\end{math} electrode array placed over the main muscle innervation zone was sampled at 2048Hz having been band-pass filtered at 10-500 Hz. 

Gradient convolution kernel compensation with an additional k-means source refinement step was implemented using the tensorflow machine learning package\cite{holobar2007multichannel}\cite{Negro2016}. As the label set was to be artificially corrupted it was important that the original be as noise-free as possible, so additional post hoc steps were taken to maximise the likelihood that the timestamps were correct. Sources were manually cleaned by examining interspike intervals and the source-to-noise ratio of each activation. An additional step of validating decomposition accuracy was implemented by comparing the sources to those found using the DEMUSE source-separation software package\cite{Holobar}\cite{holobar2007multichannel}, with source cleaning completed by a different trained operator.

\subsection{Label Set Corruption}
In experiment 1 we evaluated the ability of DeepLSAL to clean a label set corrupted by feature-dependent noise, where a label flipping probability is related to its associated features\cite{algan2020label}. In the context of source-separated HD-sEMG, this most commonly occurs as a false-positive, where a separation vector incorrectly assigns a high probability of an in-class MUAP being present when it is not, i.e. a noise class or other MU class label is flipped to the MU class of interest. To simulate this effect, we corrupted the label set by generating an artificially-noisy separation vector for each MU class; randomly-selecting 15 MUAP labels from that class and using the average of the associated extended HD-sEMG vectors to generate a linear minimum mean square error prediction on the extended HD-sEMG matrix. A two-class k-means clustering algorithm was then used to parameterise a linear threshold to find activations, creating a label set with a high degree of feature-dependent noise. Five levels of increasing difficulty were generated by taking an amount of false positives corresponding to 10/20/30/40/50\% of the number of true labels, selected at random from the set of false positives.

In experiment 2 the DeepLSAL algorithm was evaluated on class-dependent label noise, when the probability of a label flipping to another class is stable across all labels in the class\cite{algan2020label}. In HD-sEMG source separation this error generally occurs when the separation vectors are very similar, usually due to similar MUAP waveform shapes between two MU classes. This can be simulated by transferring a percentage of labels to a similar MU class. This was done by first averaging the MUAPs of each MU class and then cross-correlating these averages with the average MUAP of every other class in the recording, with 10/20/30/40\% of the class labels transferred to the class with the highest value. If labels had already been transferred to the closest class then the next closest class was selected until all classes had had label transfers. A maximum label corruption of 40\% was used to preserve the concept of a majority true and minority false class.

\subsection{DeepLSAL Pipeline and Training}
To convert the source-separated HD-sEMG signal into labelled windows, first each channel of the HD-sEMG signal was standardised by z-scoring and then cut into overlapping 80-sample wide windows at a stride of 1. Each window was then labelled by reference to the predicted source activity at the final sample of the window. This meant the bulk of windows were labelled as part of the inactive class due to the sparse nature of motor neuron spiking. Due to this serious class imbalance, each minibatch was created from the entirety of the windows labelled as containing a motor neuron spike, with an additional 256 samples from the inactive class. Each class assignment was then converted to a one-hot representation, the bulk of which had only one class active at any one time, although rarely two activations would occur simultaneously on the same time-point. As the richness of the intra-class embedding of the inactive class windows was not of any great concern, the embeddings of these windows were not used as anchor vectors when calculating the \begin{math}l_{LSAL}\end{math} component of the loss, although they were used as both \begin{math}\mathbf{x_{n}}\end{math} values and in the calculation of \begin{math}l_{cross}\end{math}. 

Embeddings were calculated with a convolutional neural network implemented using the tensorflow machine learning library in python, as seen in figure \ref{figure:2}a. Convolution steps used a 1D 3-sample wide kernel, with 32 filters and a drop-out of 0.2. 1D max-pooling was completed with 2-sample wide kernels. Each densely-connected layers had 64 neurons and a drop-out percentage of 0.5 during training. Both the convolution and densely-connected layers used ReLU activation functions. Finally the output of the last densely-connected layer was densely-connected to a bias and activation-free embedding layer of 8 neurons-wide, which was then divided by its L2 norm. This was an intentionally low-dimension embedding compared to standard DML due to the desire to avoid dimensionality issues during the clustering steps in the refitting phase. The additional matrix \begin{math}\mathbf{W}\end{math} used in the categorical cross-entropy was initialised with truncated normal noise, whilst the weights of the neural network layers was initialised by glorot uniform. The Adam optimisation algorithm at a learning rate of 0.001 was then used to train the model over 500 epochs for both cleaning and refitting stages. 

For both experiments \begin{math}k\end{math} was set to 5, \begin{math}\gamma \end{math} was set to 0.1, and \begin{math}\alpha \end{math} to 0.25 radians in both the cleaning and refitting stages. After the cleaning stage the labels in the embedding space likely to relate to specific classes were selected by a simple density-estimator. First a local scale value \begin{math}v\end{math} was estimated by finding the mean cosine similarity of the each embedding vector to its 20 nearest neighbours and taking a median of this value across all vectors. For each label the number of other labels with a cosine similarity more than \begin{math}v\end{math} was found and the label with the highest number of neighbours was selected as the centre of the cluster. All labels within a cosine similarity higher than \begin{math}v\end{math} were then added to the refitting training set. This simple approach was generally adequate for quickly finding the densest region of the embedding space, which was usually the cluster of true labels. 

After the cleaning phase the set of timestamp labels is generally free of false positives; if DeepLSAL is retrained with this new label set then it should effectively generalise to find unlabelled activations in both the current and future data (fig. \ref{figure:2}b). It should be noted that finding unlabelled activations in the current dataset is in some ways a "harder" problem than trying to generalise to completely unseen data, as these activations are included in training, but mislabelled. However these mislabelled activations are a very small proportion of the total dataset, meaning they were predicted not to impact convergence on a model with good generalisation ability, particularly a heavily regularised model such as that used for DeepLSAL. Once DeepLSAL was retrained an average embedding vector of the current MUAP timestamps was found for each class, which was then cross-correlated with the entire embedded HD-sEMG signal to generate a predicted activity (fig. \ref{figure:2}c). This activity was then timestamped by a linear threshold parameterised by a two-class k-means clustering algorithm. These labels were compared to the pre-corrupted data using the rate of agreement (RoA) metric, a percentage defined as the number of true positive matches divided by the total number of true positives, false positives and false negatives. 

\section{Results}
\subsection{Feature-dependent Label Noise}
\begin{figure*}
\centering
\includegraphics[width=0.7\textwidth]{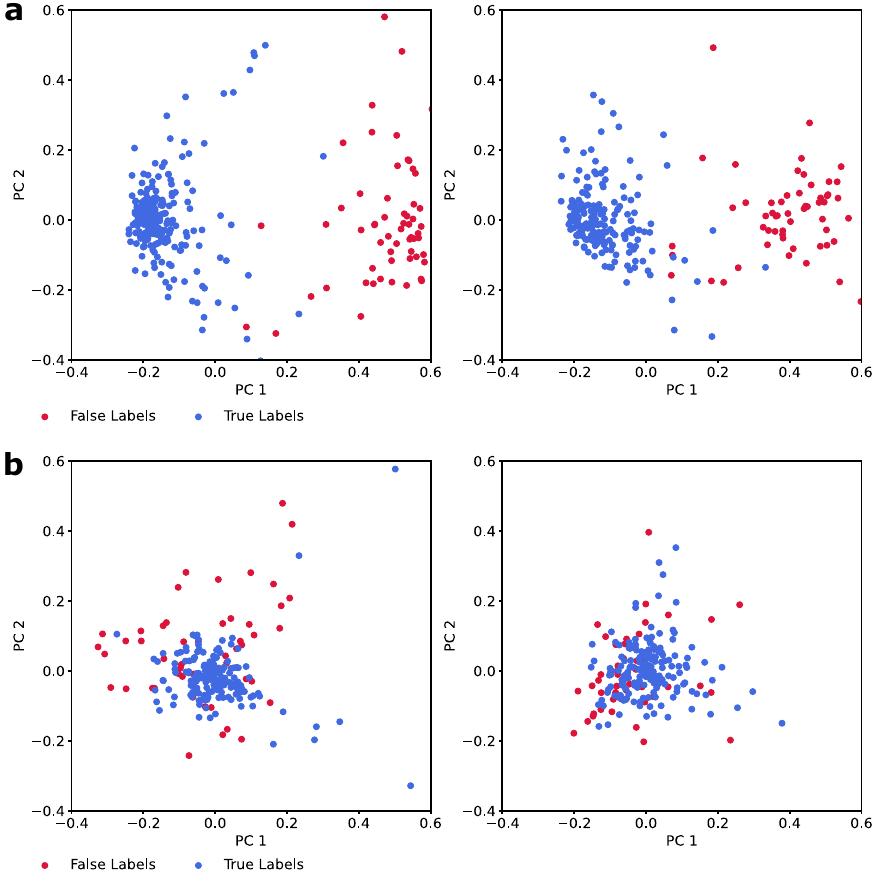}
\caption{The effect of two different sampling strategies on the embedding space for two units as shown by the first and second principle components. \textbf{a} shows the effect of locality-sensitive sampling, with a rich intra-class embedding that clearly separates the true and false embeddings. In \textbf{b} the same optimisation was run again, but with the positive and negative pairs randomly selected, leading to all intra-class embeddings contracting down to a point.}
\label{figure:2first}
\end{figure*}

\begin{figure*}
\centering
\includegraphics[width=0.7\textwidth]{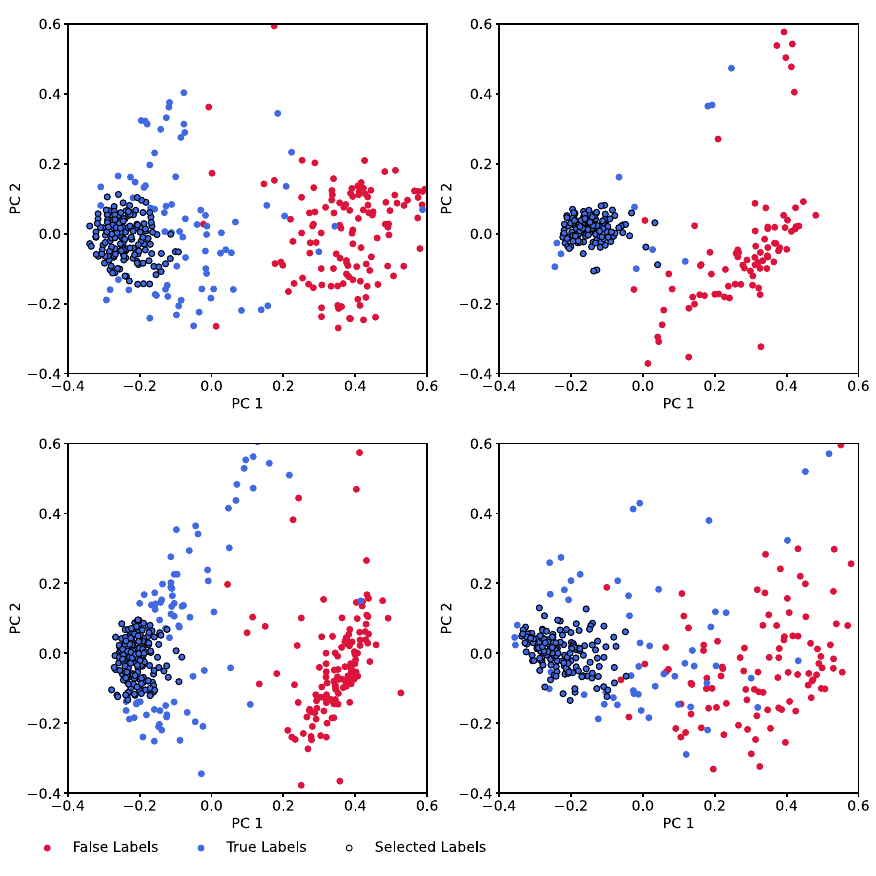}
\caption{The first and second principal components of the embedding space after the cleaning phase for all classes found in a single HD-sEMG sample, with labels initially extracted from a noisy separation filter. The samples selected by density analysis are circled, these will form the training labels for the refitting phase. The model is effective at inducing clustering of similar labels, with the area of highest density corresponding to the true labels.}
\label{figure:2a}
\end{figure*}

In experiment 1, which tested the effect of feature-dependent label noise by simulating noisy separation vectors, DeepLSAL generated an embedding space with dense clusters for each class corresponding to the true labels. Surrounding each cluster is a large sparse periphery of false labels, with no apparent structure. The efficacy of locality-sensitive sampling at preserving a rich class embedding is clear when compared to random sampling using a two-dimensional principal component space, as in figure \ref{figure:2first}. The simple density-estimator operating on the cosine similarity between embeddings could then be used to select a subset of true labels by selecting the area with the highest density (fig. \ref{figure:2a}). This algorithm was weighted to favour specificity over sensitivity, meaning almost no false labels were included in the cleaned label-set at the cost of losing a percentage of the true labels. 

\begin{figure*}
\centering
\includegraphics[width=1\textwidth]{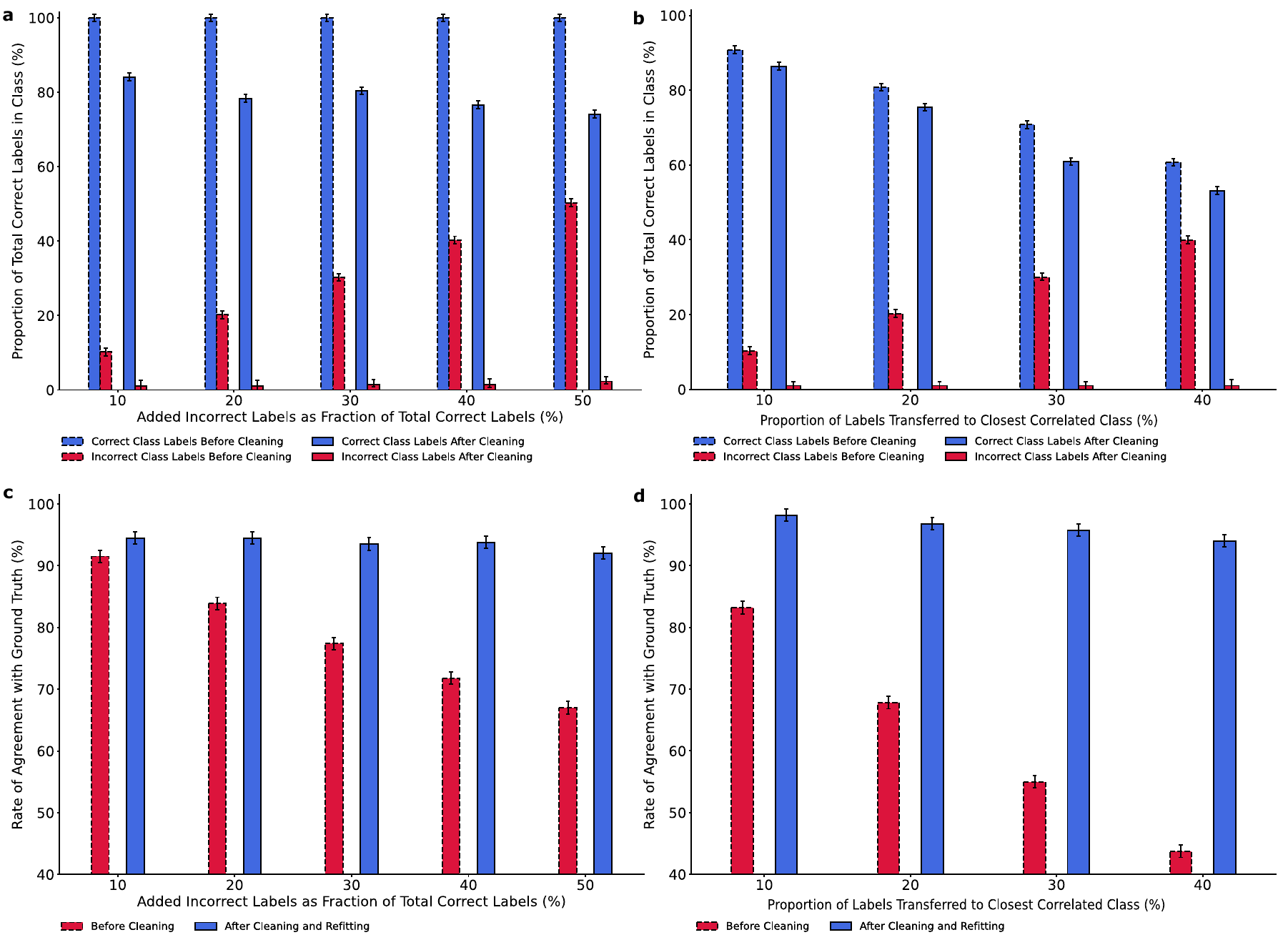}
\caption{Median label accuracy and rate of agreement plots for both experiments across all recordings, with interquartile range. \textbf{a} shows the outcome of the DeepLSAL cleaning phase for experiment 1 simulating a noisy separation filter. This leads to high number of false positives in the labels, plotted here as a fraction of the total correct labels in each class. These are almost completely removed, at the cost of losing a fraction of the true labels. A similar cleaning result was found in experiment 2 (\textbf{b}), where a proportion of labels from each class was transferred to the nearest correlated class.  Refitting recovers the bulk of these lost labels, giving good final rates of agreement with the ground truth labels. \textbf{c} shows the RoA from experiment 1 before and after cleaning and refitting, whilst \textbf{d} shows the RoA change in experiment 2. The RoA is returned to values close to pre-corruption levels.}
\label{figure:2b}
\end{figure*}

DeepLSAL generated an embedding space with utility for removing false labels even at the maximum tested value of 50\% of total correct values (fig. \ref{figure:2b}\textbf{a}), with a median post-cleaning false label retention of 1.3\% of the total correct labels in the class (IQR 0.5 - 1.9). The number of true labels lost during the cleaning process fell as the pre-cleaning percentage of false labels increased, but even at the highest false label percentage tested, a median of 74.1\% (IQR 68.9 - 80.3) of the true values were still retained (table \ref{table:1}). 
\begin{table*}
\renewcommand{\arraystretch}{1}
\caption{Combined median (interquartile range) scores for all classes across all recordings from experiment 1 for different stages of the cleaning and refitting pipeline.}
\centering
\begin{tabular}{|p{3cm}||p{3cm}||p{3cm}||p{3cm}|}
\hline
\bfseries False Labels Added as Proportion of Class  (\%) & \bfseries Starting RoA (\%) & \bfseries Remaining True Labels After Cleaning (\%) & \bfseries RoA After Cleaning and Refitting (\%) \\
\hline\hline
10 & 91.5 (91.5 – 91.7)  & 84.1 (79.6 - 87.1)  & 94.5 (91.2 – 97.2)\\
20 & 83.9 (83.8 – 83.9)  & 78.3 (74.4 - 83.9)  & 94.5 (89.8 – 97.3) \\
30 & 77.4 (77.3 – 77.5) & 80.3 (75.3 - 85.1)  & 93.5 (90.2 – 97.7) \\
40 & 71.8 (71.7 – 71.9)  & 76.6 (71.7 - 81.8) & 93.8 (87.7 – 96.3) \\
50 & 67.0 (66.9 – 67.0)  & 74.1 (68.9 - 80.3)  & 92.0 (86.7 – 96.4) \\
\hline
\end{tabular}
\label{table:1}
\end{table*}
\subsection{Class-dependent Label Noise}
\begin{figure*}
\centering
\includegraphics[width=0.7\textwidth]{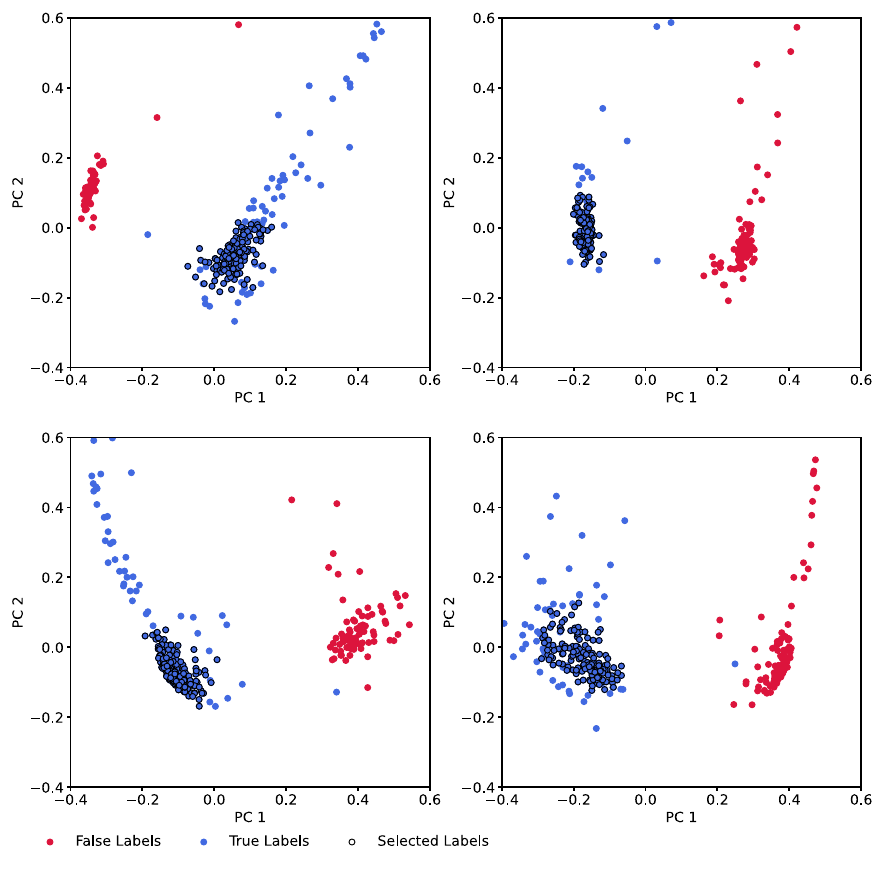}
\caption{Principal component plot of the embedding space after the cleaning phase for all classes found in a single HD-sEMG sample in experiment 2. 40\% of the labels in each class have been removed and then added to the class with the closest correlation. The samples selected automatically for the refitting phase have been circled. Unlike in experiment 1, the false labels are correlated as they come from the same class. This results in two tight clusters for both true and false labels, however they still clearly separable.}
\label{figure:2c}
\end{figure*}

In experiment 2, when labels were randomly flipped to the MU class with the closes average MUAP shape, DeepLSAL again generated embedding spaces with clear separation between true and false separation (fig. \ref{figure:2c}). However, unlike in the first experiment, the false labels formed a second distinct cluster within the embedding space, again clearly visualised in the first two principle component dimensions. As the true label cluster always had more values, it was still clearly identified by the density-estimator.  

As in experiment 1, the DeepLSAL cleaning phase was effective at removing almost all false labels (fig. \ref{figure:2b}\textbf{b}). Even at a 40\% transfer the median post-cleaning fraction of 0.0\% (0 - 0.6) of the total correct labels. As true labels were lost both to the initial transfer to other classes and to the cleaning phase, far fewer were retained in the post-cleaning dataset than in experiment 1 and would need to be recovered in the refitting stage (table \ref{table:2}). 
\subsection{Rediscovering Unlabelled Activations}
\begin{figure*}
\centering
\includegraphics[width=0.7\textwidth]{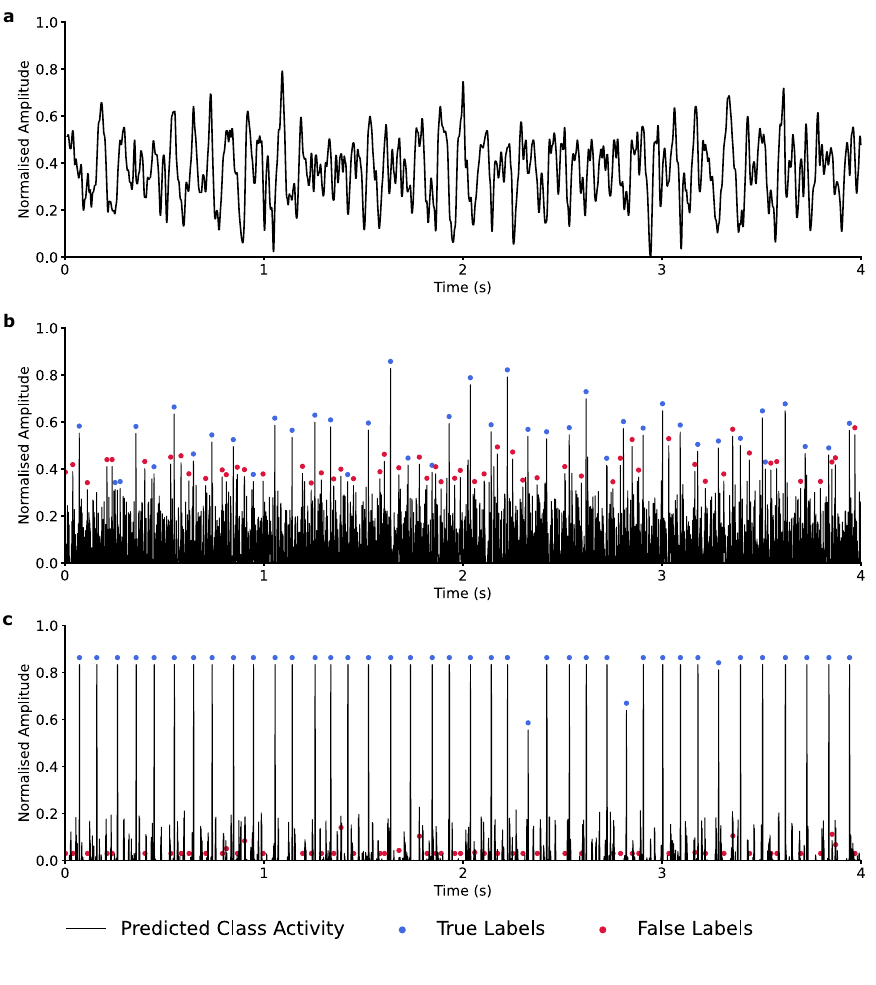}
\caption{A single channel of unprocessed HD-sEMG and the post-decomposition predicted activity of a single class before and after cleaning and refitting, with true and false labels. \textbf{a} demonstrates the degree of complex superposition inherent to sEMG signal as opposed to "cleaner" recordings such as those from intracortical sources. A linear separation filter based on an average of only 15 labels is applied to the signal to generate \textbf{b}, which is consequently extremely noisy, simulating a poorly optimised filter. A number of false positives corresponding to 50\% of the number of true class labels has been selected. After the cleaning and refitting phases the spiking motor neuron activity in \textbf{c} is clearly identifiable, whilst incorrect labels have been suppressed.}
\label{figure:2d}
\end{figure*}

An important requirement if the DeepLSAL-cleaned label set is to be useful is that the cleaning process does not overly bias against true labels that are lost at this stage, making them difficult to recover using the retained true labels. Lost labels tend to be more peripheral in the cluster, meaning their MUAP shapes are likely to be less similar to the MU class average, potentially due to superposition with a MUAP from a different class or due to a noise artefact. False negatives are also still used for training, but are labelled inappropriately, with a possibly detrimental effect of the model to generalise. To demonstrate that neither of these potential problems actually impacted training, after the cleaning stage of both experiment 1 and 2 DeepLSAL was refitted with the cleaned label sets. For both experiments the predicted activity was generally both sparse and clean, with MUAPs easily identifiable (fig. \ref{figure:2d}). These labels were compared to the original data using the rate of agreement (RoA) metric, a percentage defined as the number of true positive matches divided by the total number of true positives, false positives and false negatives. 

The RoA of the predicted MUAP labels with the original data was generally good for both experiments at every level of difficulty (tables \ref{table:1} and \ref{table:2}). In experiment 1 there was little change in RoA as difficulty increased (fig. \ref{figure:2b}\textbf{c}), suggesting that DeepLSAL is able to generalise to unseen activations. This finding was also replicated in experiment 2 (fig. \ref{figure:2b}\textbf{d}), and even when the median post-cleaning training set was just over half of the total class activations a median RoA of 94\% (86.9 - 97.7) was achieved after refitting.
\begin{table*}
\renewcommand{\arraystretch}{1}
\caption{Combined median (interquartile range) scores for all classes across all recordings from experiment 2 for different stages of the cleaning and refitting pipeline.}
\centering
\begin{tabular}{|p{3cm}||p{3cm}||p{3cm}||p{3cm}|}
\hline
\bfseries Proportion of Class Labels Transferred (\%) & \bfseries Starting RoA (\%) & \bfseries Remaining True Labels After Cleaning (\%) & \bfseries RoA After Cleaning and Refitting (\%) \\
\hline\hline
10 & 83.2 (82.1 - 83.9) & 84.1 (84.1 - 88.1) & 98.2 (95.4 – 98.8)\\
20 & 67.0 (66.1 – 69.0) & 75.4 (70.0 – 77.5) & 96.8 (93.3 – 99.0) \\
30 & 55.0 (53.1 – 56) & 60.9 (57.3 – 64.9) & 95.7 (88.9 – 98.8) \\
40 & 43.7 (42.1 - 44.9) & 53.1 (49.1 – 56.7) & 94 (86.9 – 97.7) \\
\hline
\end{tabular}
\label{table:2}
\end{table*}

\section{Discussion}
Source separation is often applied to decode the neural information embedded in neurophysiological data. This approach provides a window into the neural determinants of behaviour as well as a way to identify neural features for human machine interfacing. However, conventional source separation approaches as well as manual decompositions by expert operators often provides noisy outputs due to decoding errors.  In this study we demonstrated that a supervised DML paradigm can find self-correcting data embeddings that allow accurate cleaning of labels corrupted by common noise issues, whilst also predicting new labels not included within the training set. Importantly the model was not biased against labels not selected during the cleaning phase when refitting, which is particularly important in neurophysiological signals, such as HD-sEMG, where the shape of a subset of activations will be distorted by temporal superpositions. This ability of the model to operate on un-preprocessed signals and generalise to unseen activations is particularly relevant to the field of sEMG decomposition for prosthetic control, where a current focus is on the use of neural networks to directly identify MUAPs in raw signal and so avoid the latency involved with the current complex preprocessing pipeline that prevents online implementations\cite{Clarke2021}\cite{Wen2021}. 

Deep metric learning has a number of attractive properties over standard softmax-based binary classifiers for neurophysiological time series classification, needing fewer training examples in general and being able to adapt to new classes easily\cite{oh2016deep}. DML methods can adapt quickly to the changes in class activity commonly seen in neural systems over time, such as with dropped units in intracortical recordings or MU recruitment and derecruitment in sEMG and iEMG recordings\cite{Dunlap2020}\cite{Farina2002}. However, the focus of most implementations of DML is a high degree of inter-class separation\cite{qian2020softtriple}, rather than the need for a more descriptive intra-class embedding needed if this space is to be used for the identification of false labels. An important result of this study was that a loss function designed to operate only on samples local to each other in the embedding space can give rise to a richer intra-class distribution, avoiding the collapse to a dense point commonly seen in a more traditional triplet learning paradigm. Furthermore, we demonstrate that the utility of the embedding space generated by DeepLSAL for label cleaning is unaffected by feature- or class-dependence in the label noise. 

Whilst this study focused on source-separated HD-sEMG signal it is important to emphasise the broader applicability of this approach to any imperfectly-labelled neurophysiological time series data in which the underlying sources are repeating events, i.e. to any neural recording. Whilst the study focused specifically on action potentials, the proposed methodology could also be used for pattern recognition in bulk neurophysiological signal, supplementing recently-proposed systems of prosthetic control\cite{Pancholi2020}. Additionally, the labelling process need not be by a BSS algorithm. For example, a DeepLSAL could be applied to a dataset for which only a small component of the data has been manually labelled by an expert-operator to recover the rest of the labels accurately. In this way, the proposed approach can be viewed as a minimally supervised method for neurophysiological time series decomposition into individual cell activities.

A potential limitation of the study is the method of converting the embedding layer to a cleaned label set. One strength of the DeepLSAL algorithm was that it was usually quite simple to identify the main cluster of clean labels in the embedding space during the cleaning phase, meaning a simple density-estimator was sufficient to set a decision threshold. However, this approach tended to cut out a larger proportion of true labels than was potentially necessary, even if the refitting stage was able to rediscover those lost labels. An improved sorting methodology is even more relevant to the specific case when two classes were mixed with equal proportions, which meant that there was no way of identifying of which of the label clusters was "correct". This implies an interesting future direction for a DML-based approach is to investigate the use of adaptive k-means clustering to find distinct clusters within the error labels and hence potentially rehabilitate their averages with their origin class. Whilst it was not a focus of this study, a an additional interesting future direction might be found in improving the richness of non-activation embeddings, which may allow the identification of new unlabelled classes through clustering events. 

In summary, we have presented DeepLSAL, a deep metric learning pipeline for embedding source-separated multivariate neurophysiological time series into a dimensionally-reduced space suitable for both classification and label cleaning. Whilst the focus of the demonstration of this approach in this paper was performed on electromyographic signal decomposition, the method is broadly applicable to other neuromuscular recordings, such as intracortical or intraneural signals.

\bibliographystyle{ieeetr}
\bibliography{manuscript}

\begin{thebibliography}{10}

\bibitem{merletti2016surface}
R.~Merletti and D.~Farina, {\em Surface electromyography: physiology,
  engineering, and applications}.
\newblock John Wiley \& Sons, 2016.

\bibitem{Stark2007}
E.~Stark and M.~Abeles, ``Predicting movement from multiunit activity,'' {\em
  Journal of Neuroscience}, vol.~27, pp.~8387--8394, Aug. 2007.

\bibitem{Drebitz2019}
E.~Drebitz, B.~Schledde, A.~K. Kreiter, and D.~Wegener, ``Optimizing the yield
  of multi-unit activity by including the entire spiking activity,'' {\em
  Frontiers in Neuroscience}, vol.~13, Feb. 2019.

\bibitem{Farina2016}
D.~Farina and A.~Holobar, ``Characterization of human motor units from surface
  {EMG} decomposition,'' {\em Proceedings of the {IEEE}}, vol.~104,
  pp.~353--373, Feb. 2016.

\bibitem{Kapelner2019}
T.~Kapelner, I.~Vujaklija, N.~Jiang, F.~Negro, O.~C. Aszmann, J.~Principe, and
  D.~Farina, ``Predicting wrist kinematics from motor unit discharge timings
  for the control of active prostheses,'' {\em Journal of {NeuroEngineering}
  and Rehabilitation}, vol.~16, Apr. 2019.

\bibitem{Calvin1973}
W.~H. Calvin, ``Some simple spike separation techniques for simultaneously
  recorded neurons,'' {\em Electroencephalography and Clinical
  Neurophysiology}, vol.~34, pp.~94--96, Jan. 1973.

\bibitem{Simon1965}
W.~Simon, ``The real-time sorting of neuro-electric action potentials in
  multiple unit studies,'' {\em Electroencephalography and Clinical
  Neurophysiology}, vol.~18, pp.~192--195, Feb. 1965.

\bibitem{Rey2015}
H.~G. Rey, C.~Pedreira, and R.~Q. Quiroga, ``Past, present and future of spike
  sorting techniques,'' {\em Brain Research Bulletin}, vol.~119, pp.~106--117,
  Oct. 2015.

\bibitem{Kevric2017}
J.~Kevric and A.~Subasi, ``Comparison of signal decomposition methods in
  classification of {EEG} signals for motor-imagery {BCI} system,'' {\em
  Biomedical Signal Processing and Control}, vol.~31, pp.~398--406, Jan. 2017.

\bibitem{Negro2016}
F.~Negro, S.~Muceli, A.~M. Castronovo, A.~Holobar, and D.~Farina,
  ``Multi-channel intramuscular and surface {EMG} decomposition by convolutive
  blind source separation,'' {\em Journal of Neural Engineering}, vol.~13,
  p.~026027, Feb. 2016.

\bibitem{Pachitariu2016}
M.~Pachitariu, N.~Steinmetz, S.~Kadir, M.~Carandini, and H.~K. D., ``Kilosort:
  realtime spike-sorting for extracellular electrophysiology with hundreds of
  channels,'' {\em bioRxiv preprint}, June 2016.

\bibitem{Steinmetz2017}
J.~J. Jun, N.~A. Steinmetz, J.~H. Siegle, D.~J. Denman, M.~Bauza, B.~Barbarits,
  A.~K. Lee, C.~A. Anastassiou, A.~Andrei, {\c{C}}.~Ayd{\i}n, M.~Barbic, T.~J.
  Blanche, V.~Bonin, J.~Couto, B.~Dutta, S.~L. Gratiy, D.~A. Gutnisky,
  M.~H\"{a}usser, B.~Karsh, P.~Ledochowitsch, C.~M. Lopez, C.~Mitelut, S.~Musa,
  M.~Okun, M.~Pachitariu, J.~Putzeys, P.~D. Rich, C.~Rossant, W.~lung Sun,
  K.~Svoboda, M.~Carandini, K.~D. Harris, C.~Koch, J.~O'Keefe, and T.~D.
  Harris, ``Fully integrated silicon probes for high-density recording of
  neural activity,'' {\em Nature}, vol.~551, pp.~232--236, Nov. 2017.

\bibitem{Muceli2019}
S.~Muceli, W.~Poppendieck, K.-P. Hoffmann, S.~Dosen, J.~Benito-Le{\'{o}}n,
  F.~O. Barroso, J.~L. Pons, and D.~Farina, ``A thin-film multichannel
  electrode for muscle recording and stimulation in neuroprosthetics
  applications,'' {\em Journal of Neural Engineering}, vol.~16, p.~026035, Feb.
  2019.

\bibitem{DelVecchio2019b}
A.~D. Vecchio and D.~Farina, ``Interfacing the neural output of the spinal
  cord: robust and reliable longitudinal identification of motor neurons in
  humans,'' {\em Journal of Neural Engineering}, Oct. 2019.

\bibitem{Clarke2021}
A.~K. Clarke, S.~F. Atashzar, A.~D. Vecchio, D.~Barsakcioglu, S.~Muceli,
  P.~Bentley, F.~Urh, A.~Holobar, and D.~Farina, ``Deep learning for robust
  decomposition of high-density surface {EMG} signals,'' {\em {IEEE}
  Transactions on Biomedical Engineering}, vol.~68, pp.~526--534, Feb. 2021.

\bibitem{Wen2021}
Y.~Wen, S.~Avrillon, J.~C. Hernandez-Pavon, S.~J. Kim, F.~Hug, and J.~L. Pons,
  ``A convolutional neural network to identify motor units from high-density
  surface electromyography signals in real time,'' {\em Journal of Neural
  Engineering}, Mar. 2021.

\bibitem{algan2020label}
G.~Algan and I.~Ulusoy, ``Label noise types and their effects on deep
  learning,'' {\em arXiv preprint arXiv:2003.10471}, 2020.

\bibitem{Kumar2020}
R.~I. Kumar, M.~M. Mallette, S.~S. Cheung, D.~W. Stashuk, and D.~A. Gabriel,
  ``A method for editing motor unit potential trains obtained by decomposition
  of surface electromyographic signals,'' {\em Journal of Electromyography and
  Kinesiology}, vol.~50, p.~102383, Feb. 2020.

\bibitem{McGill2005}
K.~C. McGill, Z.~C. Lateva, and H.~R. Marateb, ``{EMGLAB}: An interactive {EMG}
  decomposition program,'' {\em Journal of Neuroscience Methods}, vol.~149,
  pp.~121--133, Dec. 2005.

\bibitem{Hug2021}
F.~Hug, S.~Avrillon, A.~D. Vecchio, A.~Casolo, J.~Ibanez, S.~Nuccio,
  J.~Rossato, A.~Holobar, and D.~Farina, ``Analysis of motor unit spike trains
  estimated from high-density surface electromyography is highly reliable
  across operators,'' {\em bioRxiv preprint}, Feb. 2021.

\bibitem{Carlson2019}
D.~Carlson and L.~Carin, ``Continuing progress of spike sorting in the era of
  big data,'' {\em Current Opinion in Neurobiology}, vol.~55, pp.~90--96, Apr.
  2019.

\bibitem{Yger2018}
P.~Yger, G.~L. Spampinato, E.~Esposito, B.~Lefebvre, S.~Deny, C.~Gardella,
  M.~Stimberg, F.~Jetter, G.~Zeck, S.~Picaud, J.~Duebel, and O.~Marre, ``A
  spike sorting toolbox for up to thousands of electrodes validated with ground
  truth recordings in vitro and in vivo,'' {\em {eLife}}, vol.~7, Mar. 2018.

\bibitem{song2020learning}
H.~Song, M.~Kim, D.~Park, and J.-G. Lee, ``Learning from noisy labels with deep
  neural networks: A survey,'' {\em arXiv preprint}, 2020.

\bibitem{Karimi2019}
D.~Karimi, H.~Dou, S.~K. Warfield, and A.~Gholipour, ``Deep learning with noisy
  labels: exploring techniques and remedies in medical image analysis,'' {\em
  CoRR}, vol.~abs/1912.02911, 2019.

\bibitem{Frenay2014}
B.~Frenay and M.~Verleysen, ``Classification in the presence of label noise: A
  survey,'' {\em {IEEE} Transactions on Neural Networks and Learning Systems},
  vol.~25, pp.~845--869, May 2014.

\bibitem{VeitACKGB17}
A.~Veit, N.~Alldrin, G.~Chechik, I.~Krasin, A.~Gupta, and S.~J. Belongie,
  ``Learning from noisy large-scale datasets with minimal supervision,'' {\em
  CoRR}, vol.~abs/1701.01619, 2017.

\bibitem{DBLP:Lee}
K.~Lee, X.~He, L.~Zhang, and L.~Yang, ``Cleannet: Transfer learning for
  scalable image classifier training with label noise,'' {\em CoRR},
  vol.~abs/1711.07131, 2017.

\bibitem{DBLP:Han}
J.~Han, P.~Luo, and X.~Wang, ``Deep self-learning from noisy labels,'' {\em
  CoRR}, vol.~abs/1908.02160, 2019.

\bibitem{Chopra}
S.~Chopra, R.~Hadsell, and Y.~LeCun, ``Learning a similarity metric
  discriminatively, with application to face verification,'' in {\em 2005
  {IEEE} Computer Society Conference on Computer Vision and Pattern Recognition
  ({CVPR}{\textquotesingle}05)}, {IEEE}, 2005.

\bibitem{wang2017deep}
J.~Wang, F.~Zhou, S.~Wen, X.~Liu, and Y.~Lin, ``Deep metric learning with
  angular loss,'' 2017.

\bibitem{Schroff_2015_CVPR}
F.~Schroff, D.~Kalenichenko, and J.~Philbin, ``Facenet: A unified embedding for
  face recognition and clustering,'' in {\em Proceedings of the IEEE Conference
  on Computer Vision and Pattern Recognition (CVPR)}, June 2015.

\bibitem{HermansBL17}
A.~Hermans, L.~Beyer, and B.~Leibe, ``In defense of the triplet loss for person
  re-identification,'' {\em CoRR}, vol.~abs/1703.07737, 2017.

\bibitem{NIPS2016_Sohn}
K.~Sohn, ``Improved deep metric learning with multi-class n-pair loss
  objective,'' in {\em Advances in Neural Information Processing Systems}
  (D.~Lee, M.~Sugiyama, U.~Luxburg, I.~Guyon, and R.~Garnett, eds.), vol.~29,
  Curran Associates, Inc., 2016.

\bibitem{Wu_2017_ICCV}
C.-Y. Wu, R.~Manmatha, A.~J. Smola, and P.~Krahenbuhl, ``Sampling matters in
  deep embedding learning,'' in {\em Proceedings of the IEEE International
  Conference on Computer Vision (ICCV)}, Oct 2017.

\bibitem{rippel2016metric}
O.~Rippel, M.~Paluri, P.~Dollar, and L.~Bourdev, ``Metric learning with
  adaptive density discrimination,'' 2016.

\bibitem{Wang_2019_CVPR}
X.~Wang, Y.~Hua, E.~Kodirov, G.~Hu, R.~Garnier, and N.~M. Robertson, ``Ranked
  list loss for deep metric learning,'' in {\em Proceedings of the IEEE/CVF
  Conference on Computer Vision and Pattern Recognition (CVPR)}, June 2019.

\bibitem{Xu2021}
F.~Xu, B.~Ma, H.~Chang, and S.~Shan, ``{PRDP}: Person reidentification with
  dirty and poor data,'' {\em {IEEE} Transactions on Cybernetics}, pp.~1--13,
  2021.

\bibitem{fan2017learning}
Y.~Fan, S.~Lyu, Y.~Ying, and B.-G. Hu, ``Learning with average top-k loss,''
  {\em arXiv preprint arXiv:1705.08826}, 2017.

\bibitem{holobar2007multichannel}
A.~Holobar and D.~Zazula, ``Multichannel blind source separation using
  convolution kernel compensation,'' {\em IEEE Transactions on Signal
  Processing}, vol.~55, no.~9, pp.~4487--4496, 2007.

\bibitem{Farina2016a}
D.~Farina, F.~Negro, S.~Muceli, and R.~M. Enoka, ``Principles of motor unit
  physiology evolve with advances in technology,'' {\em Physiology}, vol.~31,
  pp.~83--94, Mar. 2016.

\bibitem{Ji2020}
S.~Ji, Z.~Zhang, S.~Ying, L.~Wang, X.~Zhao, and Y.~Gao, ``Kullback-leibler
  divergence metric learning,'' {\em {IEEE} Transactions on Cybernetics},
  pp.~1--12, 2020.

\bibitem{ShazeerMMDLHD17}
N.~Shazeer, A.~Mirhoseini, K.~Maziarz, A.~Davis, Q.~V. Le, G.~E. Hinton, and
  J.~Dean, ``Outrageously large neural networks: The sparsely-gated
  mixture-of-experts layer,'' {\em CoRR}, vol.~abs/1701.06538, 2017.

\bibitem{Holobar}
A.~Holobar and D.~Zazula, ``Gradient convolution kernel compensation applied to
  surface electromyograms,'' in {\em Independent Component Analysis and Signal
  Separation}, pp.~617--624, Springer Berlin Heidelberg, 2007.

\bibitem{holobar2010experimental}
A.~Holobar, M.~A. Minetto, A.~Botter, F.~Negro, and D.~Farina, ``Experimental
  analysis of accuracy in the identification of motor unit spike trains from
  high-density surface emg,'' {\em IEEE Transactions on Neural Systems and
  Rehabilitation Engineering}, vol.~18, no.~3, pp.~221--229, 2010.

\bibitem{oh2016deep}
H.~Oh~Song, Y.~Xiang, S.~Jegelka, and S.~Savarese, ``Deep metric learning via
  lifted structured feature embedding,'' in {\em Proceedings of the IEEE
  conference on computer vision and pattern recognition}, pp.~4004--4012, 2016.

\bibitem{Dunlap2020}
C.~F. Dunlap, S.~C. Colachis, E.~C. Meyers, M.~A. Bockbrader, and D.~A.
  Friedenberg, ``Classifying intracortical brain-machine interface signal
  disruptions based on system performance and applicable compensatory
  strategies: A review,'' {\em Frontiers in Neurorobotics}, vol.~14, Oct. 2020.

\bibitem{Farina2002}
D.~Farina, M.~Fosci, and R.~Merletti, ``Motor unit recruitment strategies
  investigated by surface {EMG} variables,'' {\em Journal of Applied
  Physiology}, vol.~92, pp.~235--247, Jan. 2002.

\bibitem{qian2020softtriple}
Q.~Qian, L.~Shang, B.~Sun, J.~Hu, H.~Li, and R.~Jin, ``Softtriple loss: Deep
  metric learning without triplet sampling,'' 2020.

\bibitem{Pancholi2020}
S.~Pancholi and A.~M. Joshi, ``Advanced energy kernel-based feature extraction
  scheme for improved {EMG}-{PR}-based prosthesis control against force
  variation,'' {\em {IEEE} Transactions on Cybernetics}, pp.~1--10, 2020.

\end{thebibliography}

\end{document}